\documentclass[conference]{IEEEtran}
\IEEEoverridecommandlockouts

\usepackage{cite}
\usepackage{amsmath,amssymb,amsfonts}
\usepackage{algpseudocode}
\usepackage{algorithm}
\usepackage{graphicx}
\usepackage{textcomp}
\usepackage{xcolor}
\usepackage{multirow}
 \usepackage{hyperref}
 \hypersetup{
     colorlinks=true,
     linkcolor=blue,
     filecolor=blue,
     citecolor = blue,      
     urlcolor=cyan,
     }

\usepackage{booktabs}
\def\BibTeX{{\rm B\kern-.05em{\sc i\kern-.025em b}\kern-.08em
    T\kern-.1667em\lower.7ex\hbox{E}\kern-.125emX}}
\begin{document}

\title{LEDRO: \underline{L}LM-\underline{E}nhanced \underline{D}esign Space \underline{R}eduction and \underline{O}ptimization for Analog Circuits\\
{\footnotesize }
\thanks{The project was funded by the MIT-IBM Watson AI Lab.}
}

\author{\IEEEauthorblockN{Dimple Vijay Kochar\textsuperscript{1}, Hanrui Wang\textsuperscript{1}, Anantha P. Chandrakasan\textsuperscript{1}, Xin Zhang\textsuperscript{2,3} }
\IEEEauthorblockA{\textit{\textsuperscript{1} Electrical Engineering and Computer Science, Massachusetts Institute of Technology, Cambridge, MA} \\
\textit{\textsuperscript{2} MIT-IBM Watson AI Lab, Cambridge, MA}\\
\textit{\textsuperscript{3} IBM T. J. Watson Research Center, Yorktown Heights, NY}\\
Emails: dkochar@mit.edu, xzhang@us.ibm.com}
}

\maketitle

\begin{abstract}
Traditional approaches for designing analog circuits are time-consuming and require significant human expertise.
Existing automation efforts using methods like Bayesian Optimization (BO) and Reinforcement Learning (RL) are sub-optimal and costly to generalize across different topologies and technology nodes.
In our work, we introduce a novel approach, LEDRO, utilizing Large Language Models (LLMs) in conjunction with optimization techniques to iteratively refine the design space for analog circuit sizing.
LEDRO is highly generalizable compared to other RL and BO baselines, eliminating the need for design annotation or model training for different topologies or technology nodes.
We conduct a comprehensive evaluation of our proposed framework and baseline on 22 different Op-Amp topologies across four FinFET technology nodes.
Results demonstrate the superior performance of LEDRO as it outperforms our best baseline by an average of 13\% FoM improvement with 2.15$\times$ speed-up on low-complexity Op-Amps and 48\% FoM improvement with 1.7$\times$ speed-up on high-complexity Op-Amps.
This highlights LEDRO's effective performance, efficiency, and generalizability.
\end{abstract}


\section{Introduction}
Designing analog circuits involves various subtasks like topology selection, transistor sizing, layout, and verification while optimizing power efficiency, maximizing performance, and reducing the area to meet the target specifications.
These subtasks are complex, requiring time-consuming simulations along with human expertise \cite{old}.
Minor topological modifications, technology node transfer, and advanced technologies such as FinFETs further increase this human effort.
Thus, there is an increasing need to automate the process of analog circuit design.

Recent advancements have introduced automation in operational amplifier (Op-Amp) design by reformulating the sizing and biasing of transistors as an optimization problem \cite{bo}.
Previous works have explored optimization techniques such as Bayesian Optimization (BO) \cite{bo, turbo1}, Reinforcement Learning (RL) \cite{gcnrl, autockt}, and genetic algorithms \cite{genetic, genetic1} to solve this problem.
However, these approaches treat the circuit design space as a black box, solving without any domain-specific analog design knowledge that an expert analog designer might have, which can lead to sub-optimal solutions.
For example, these approaches often overlook the regions of operation or key transistor biasing points.
Also, these approaches are costly and not generalizable since they often require retraining for each circuit topology and technology node \cite{gcnrl, autockt}.

\begin{figure*}[t]
    \centering
    \includegraphics[width=0.9\textwidth]{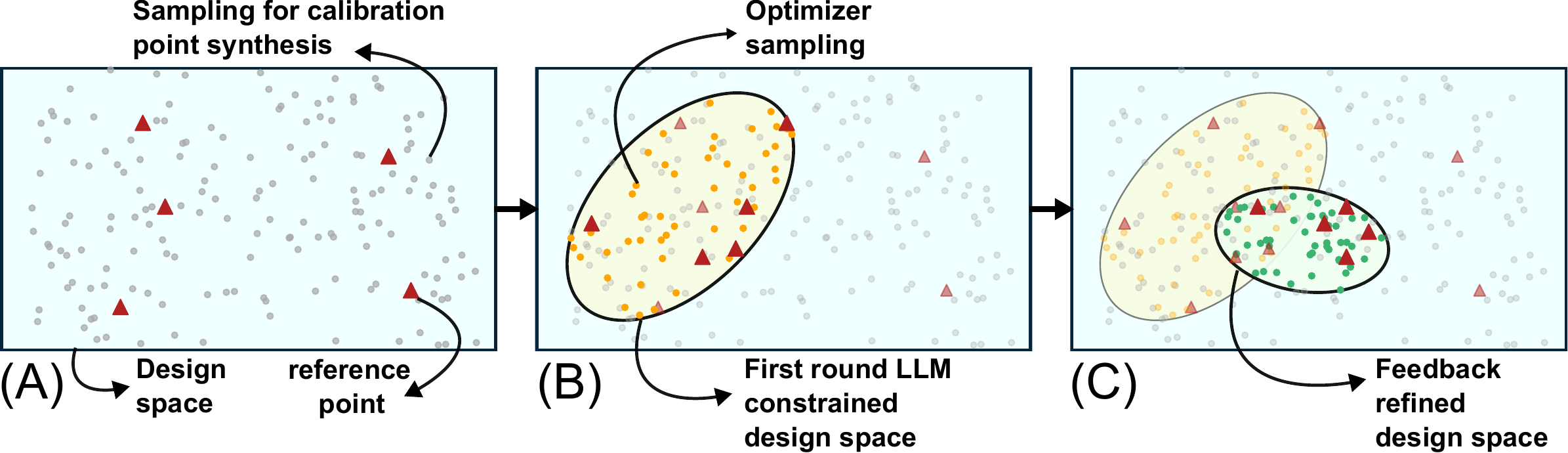}
    \caption{Illustration of LLM-enhanced design space reduction and optimization (LEDRO).}
    \label{fig1}
    \vspace{-10pt}
\end{figure*}

Recent works explore utilizing LLMs for analog circuit design \cite{abl1, abl2, ab1, ab2, ab3, ab4}, but most of them do not focus on transistor level sizing optimization. Refs. \cite{abl1, abl2} utilize LLMs for analog layout design, \cite{ab1} uses agentic LLMs to automate design process, and \cite{ab2, ab4} prioritize topology and correct functionality of analog circuits. Ref. \cite{ab3} focuses on transistor sizing optimization by expecting LLMs to directly provide the optimized sizing value, which can be challenging to scale to larger search spaces. 
To overcome these shortcomings, we introduce a novel framework - LLM-Enhanced Design Space Reduction and Optimization (LEDRO) - that synergizes the mathematical reasoning from optimization techniques with the circuit knowledge of LLMs.
Fundamentally, LEDRO enhances design space exploration by iteratively leveraging LLMs to choose a refined search region (instead of direct points) and optimization techniques to find high-performing circuits in the chosen region. 
Fig.~\ref{fig1} provides a high-level illustration of LEDRO.
To ensure LEDRO's generalizability across technology nodes and topologies, we perform calibration point synthesis using optimization techniques to provide reference examples to the LLM to generate refined regions without requiring any circuit-specific training.
To further improve LLM outputs, we provide LLM with its knowledge-based self-reflection and simulation-based optimization feedback.
In our work, we utilize LLaMa3-70B \cite{llama3} as the LLM and TuRBO \cite{turbo} as the optimization technique for LEDRO.



We compare our framework against a state-of-the-art actor-critic RL framework \cite{autockt, RL1, RL2} and a pure TuRBO approach \cite{turbo}.
To benchmark the generalizability of our framework, we evaluate it on one of the most comprehensive setups of 88 different circuits comprising a broad range of 22 Op-Amp topologies across 4 FinFET technology nodes.
We divide the topologies into low and high-complexity Op-Amps for finer comparisons.
For evaluation, we utilize the Figure of Merit (FoM), an objective function of the normalized, weighted sum of the gain, unity gain bandwidth, phase margin, and supply current.
Experiments reveal the broad generalizability of LEDRO with an average FoM improvement of 13\% with 2.15$\times$ speed-up for the low-complexity Op-Amps over our best baseline and 48\% FoM improvement with a 1.7$\times$ speed-up for the high-complexity ones.
Thus, we show that LEDRO leverages LLMs to refine the search space to more relevant ranges to ensure better and more efficient designs.
In addition, LEDRO offers wide versatility as it can be coupled with any optimization method as a plug-and-play approach.

Overall, we summarize our key contributions as follows: 
\begin{itemize}
\item We introduce a novel framework, LEDRO, which conducts \textit{refined-search space exploration} by combining LLMs with optimization techniques, becoming one of the first works to utilize LLMs for analog circuit sizing.
\item We demonstrate the generalizability of our framework on an \textit{extensive experimental setup} comprising 22 Op-Amp topologies across four different technology nodes, particularly in FinFET technology 
\item We showcase the \textit{superior performance and speed-up} of LEDRO on our benchmarks compared to state-of-the-art RL and standalone TuRBO algorithms, establishing LEDRO as a highly effective and efficient framework for analog circuit design.
\end{itemize}


The rest of this paper is organized as follows. Section \ref{sec:methods} introduces the proposed LEDRO method with a case study example. Section \ref{sec:experiments} provides experiment results on a set of diverse circuits and Section \ref{sec:conclusion} concludes the paper. The code is available at \url{https://github.com/dimplekochar/LEDRO}.



\section{Methods}
\label{sec:methods}
\begin{figure}[t]
    \centering
    \includegraphics[scale=0.34]{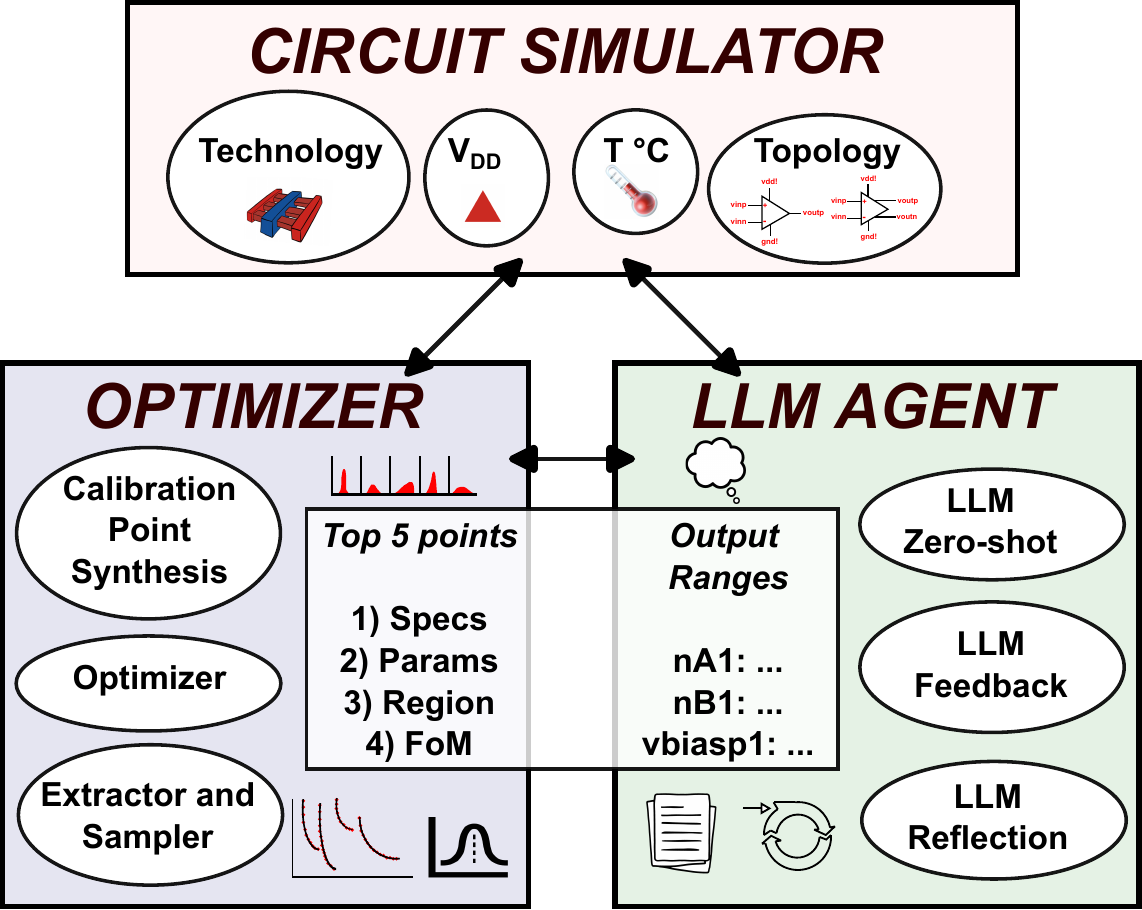}
    \caption{Block-diagram of proposed framework.}
    \label{fig1-1}
    \vspace{-10pt}
\end{figure}

Our proposed framework LEDRO utilizes a Large Language Model (LLM) in conjunction with an optimization module to more efficiently explore the search space for analog design.
A circuit simulator simulates the created analog designs and estimates the performance.
LEDRO continually explores and improves the design by running these components in iteration for a fixed set of rounds.
In our work, we utilize LLaMa3-70B \cite{llama3} as the LLM and TuRBO \cite{turbo} as our optimizer. For LLaMa3-70B, we utilize greedy decoding with temperature of 0.8, and maximum generation length of 1000 tokens. In total, we run LEDRO for ten iterations.
We provide a high-level block diagram of LEDRO in Fig.~\ref{fig1-1} and a case study in \ref{sec:case-study}.


\subsection{Calibration Point Synthesis for LLM}
\label{sec:calibration-point-synthesis}

Directly identifying optimal points or regions is a non-trivial task for LLMs since different topologies across technology nodes can have highly diverse optimal points in the complex design space.
Instead, to effectively utilize LLMs, we create a \textit{calibrated initial prompt} which aids the LLM to understand and navigate the search space better.
Specifically, we include some reference designs in this prompt that provide this search space calibration and understanding.

To provide these reference designs, we first sample 200 points by running the optimizer over the entire search space as allowed by the technology node.
Next, we extract ``good" points from these sampled points by filtering the design configurations for which the amplifier gain exceeds a threshold value.
We set the threshold to a low 0 dB so that these ``good" points are easy to find.
In rare cases where no ``good" points are found, the optimization can be run longer, or the threshold value can be lowered (although we did not face such an issue with our experiments).
Finally, we rank these filtered ``good" points on their Figure of Merit (described in the next paragraph) and add the top five designs to our \textit{calibrated initial prompt} as the reference designs, as shown in Fig~\ref{fig1}(A).

\noindent \textbf{Figure of Merit:}
To quantitatively assess the configurations, similar to \cite{autockt, marl, tbss}, we define our Figure of Merit (FoM) by an objective function utilizing the specifications $\mathcal{S}$ of gain ($G$), unity gain bandwidth ($f$), supply current ($I$), and phase margin ($PM$).
First we normalize each specification $s\in\mathcal{S}$ using its corresponding user-defined boundary value $s_{bound}\in\mathcal{S}_{bound}$ by the normalizing function $\phi$ as
$$
    \phi(s, s_{bound}) = \frac{s - s_{bound}}{s + s_{bound}}
$$
Thus, if $s$ = $s_{bound}$, the normalized value is capped at 0; else $\phi(s, s_{bound}) < 0$.
Such capping and normalization ensures the optimization of all specifications rather than the hyper-optimization of a single specification.

We formulate the FoM objective function $\mathcal{V}$ as the weighted sum of the normalized specifications as
$$
    \mathcal{V}(\mathcal{S}, \mathcal{S}_{bound}; w) = \sum_{(s,s_{bound})\in(\mathcal{S},\mathcal{S}_{bound})} w_s . \phi(s,s_{bound})
$$
where $w_s$ is the weight assigned for specification $s$.
Owing to the capping in the normalization, the maximum achievable FoM value is 0, with higher FoMs being more desirable.

\subsection{LLM: First Round Prompt}
\label{sec:llmfrp}
The design search space for amplifiers is high dimensional and complex since each transistor can have a wide array of possible configurations in terms of length, number of fins, and biasing conditions \cite{ann}.
LEDRO leverages the pre-trained knowledge and reasoning capabilities of LLMs to reduce and refine this design search space, simplifying exploration for subsequent optimizations (as shown in Fig~\ref{fig1}(B)).
More technically, given the input prompt with the calibration points, LLMs are required to output this refined search space region.

\noindent \textbf{Circuit Representation:}
To describe the circuit to LLMs in natural language, we represent it by its netlist.
The netlist provides a detailed description of the connectivity and component specifications in the circuit.
Our studies show that LLMs can parse and interpret the circuit intricacies using the netlist.

\begin{figure}[t]
\centerline{\includegraphics[scale=0.41]{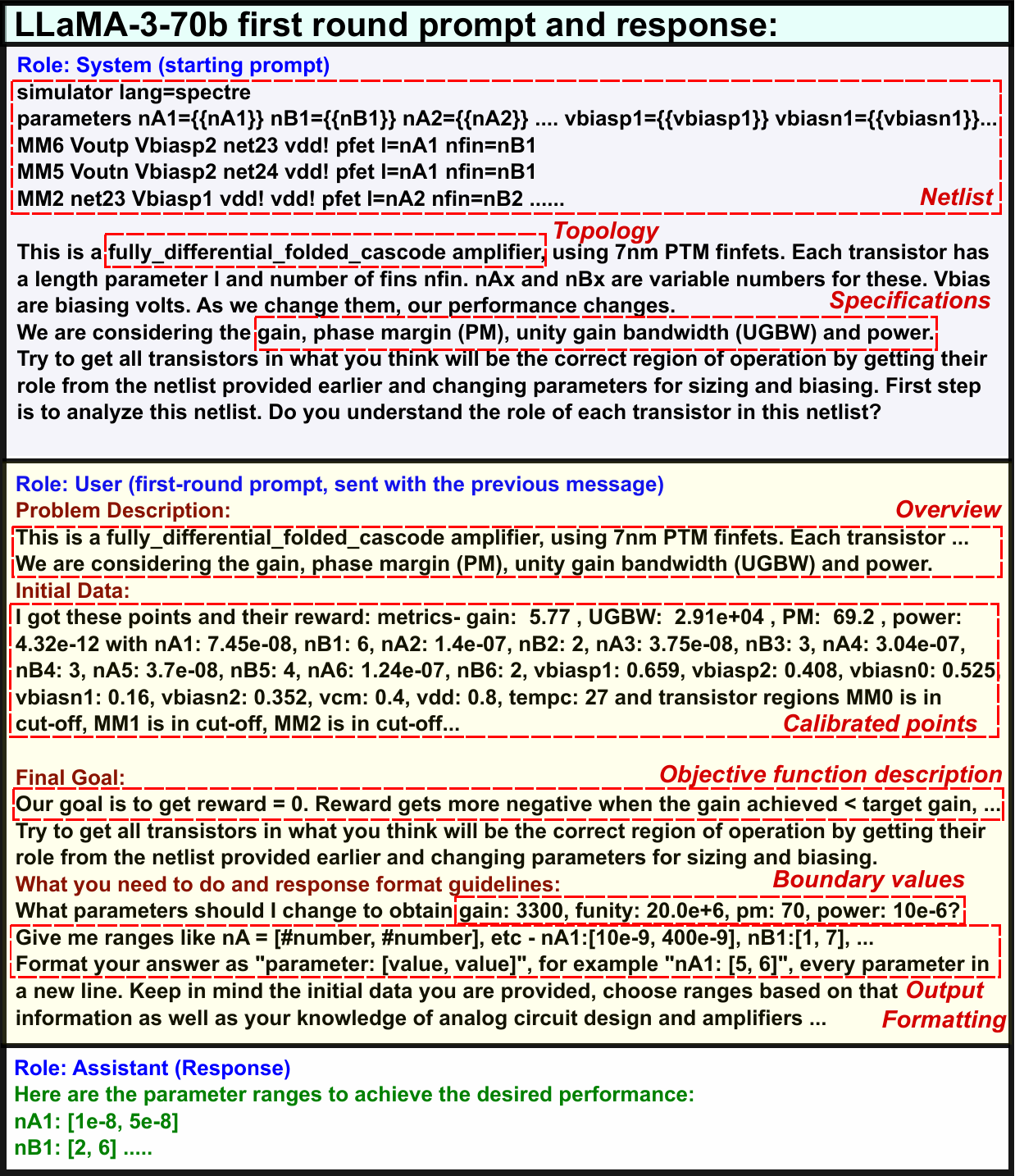}}
\vspace{-10pt}
\caption{First round prompt and response for folded cascode amplifier.}
\label{fig7}
\vspace{-10pt}
\end{figure}

\noindent \textbf{Prompt Setup:}
The LLM prompt comprises the system prompt for high-level instructions and the user prompt for specific inputs and queries.
We provide our problem description, which includes the circuit's netlist, amplifier topology, and the specifications in the system prompt.
This ensures that these details serve as a continuous reference to the LLM.

For the user prompt for the first round, we provide the problem overview and the design objectives and introduce the initial set of calibrated points (as described in \ref{sec:calibration-point-synthesis}).
Each calibrated point is characterized by the specifications ($G$, $PM$, etc), the transistor parameters (number of fins, biasing, etc.), the FoM $\mathcal{V}$, and each transistor's region of operation.
We also provide the specification boundary values and the verbalized FoM objective function.
Finally, we add instructions prompting the LLM to select and appropriately reduce the design search space, along with output formatting instructions.
Detailed information about the transistor operation (cut-off, subthreshold, saturation, or linear region) and its self-knowledge allows the LLM to make informed decisions and efficiently provide the refined search space region.
We provide an illustration of this prompt in Fig~\ref{fig7}.

\noindent \textbf{LLM to Optimization:}
The LLM-generated refined search region is then provided back to the optimizer (in our case, TuRBO).
We sample 100 points in this reduced space using the optimizer and provide corresponding feedback (discussed in~\ref{sec:llmfrp2}) to the LLM in an iterative fashion.
The refined search space from LLMs is the defining feature of LEDRO as it prioritizes regional exploitation by significantly reducing the optimization complexity over a wide-range exploration on the entire search space.
This subsequently accelerates the identification of high-performing designs and uncovers innovative configurations that might have remained undiscovered with only the optimizer due to higher complexity.

\begin{figure}[t]
\centerline{\includegraphics[scale=0.41]{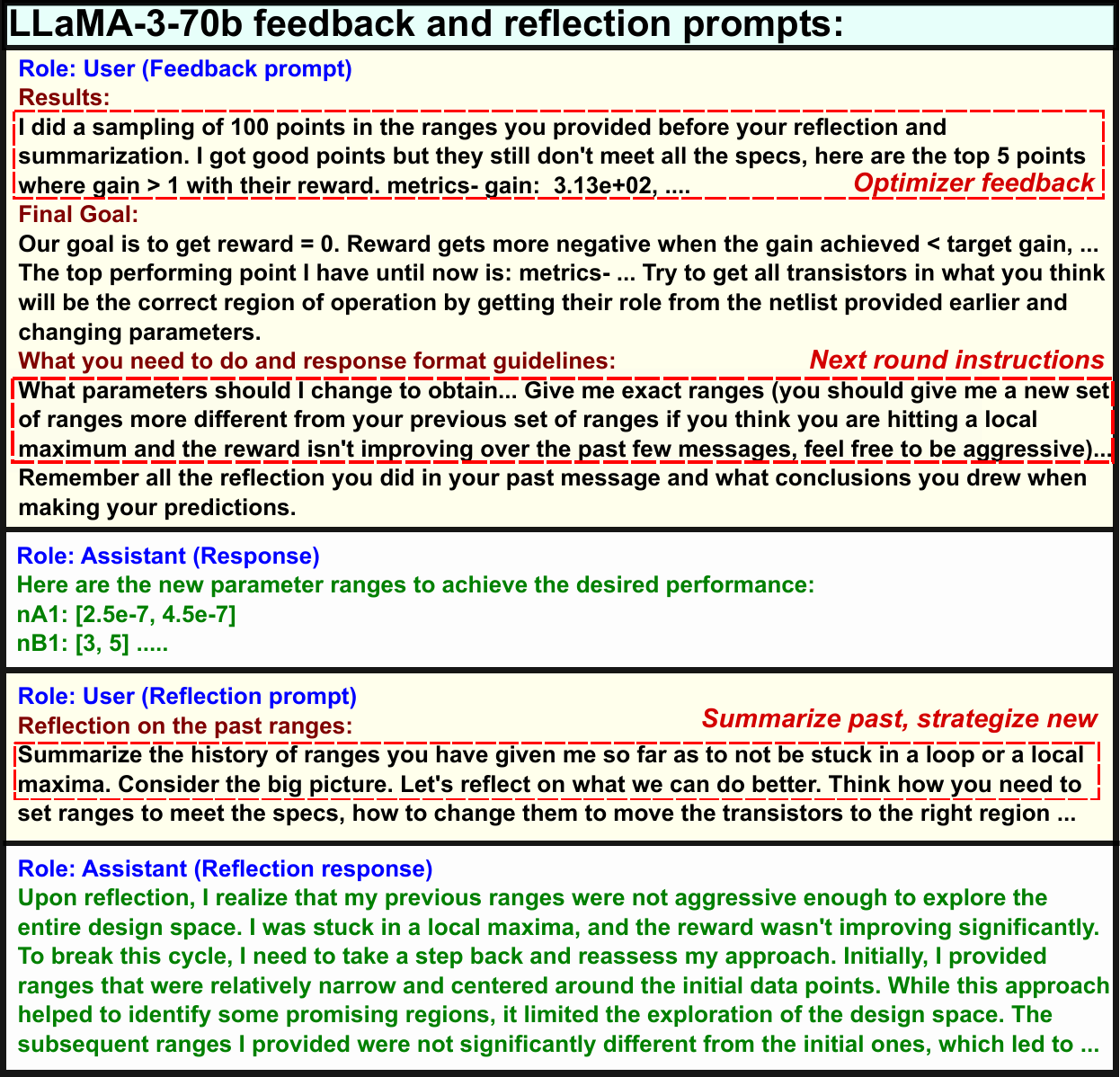}}
\vspace{-10pt}
\caption{Further feedback and reflection prompts and responses for further rounds for a folded cascode amplifier.}
\label{fig8}
\end{figure}

\subsection{LLM: Feedback and Reflection for Further Rounds}
\label{sec:llmfrp2}
For subsequent rounds, we remove the problem description from the LLM prompt and instead add the optimizer feedback and the self-reflection for effective and better-informed search space reduction, as shown in Fig~\ref{fig1}(C).
The optimizer feedback can be of two forms based on the optimizer's performance within the LLM-provided design space from the last round.
If the optimizer performs well - characterized by achieving more than five ``good" points along with an increased FoM relative to previous rounds - we provide these new points back to the LLM, acknowledging its effective performance.
Contrarily, if the optimizer performs poorly, we provide this feedback to the LLM and ask it to explore a new and different search region.

As part of self-reflection, we prompt the LLM to reflect on its past ranges and strategize what to do next.
This self-reflection is inspired by concepts outlined in \cite{reflex}, where we prompt the LLM agent to summarize previous design ranges and look back on its past strategies.
Overall, we find that the optimization feedback and self-reflection encourage the LLM to dynamically adjust its exploration strategy and efficiently enhance the refined region. 
We provide an illustration of the feedback and reflection prompts in Fig~\ref{fig8}.

\begin{figure}[t]
\centerline{\includegraphics[scale=0.23]{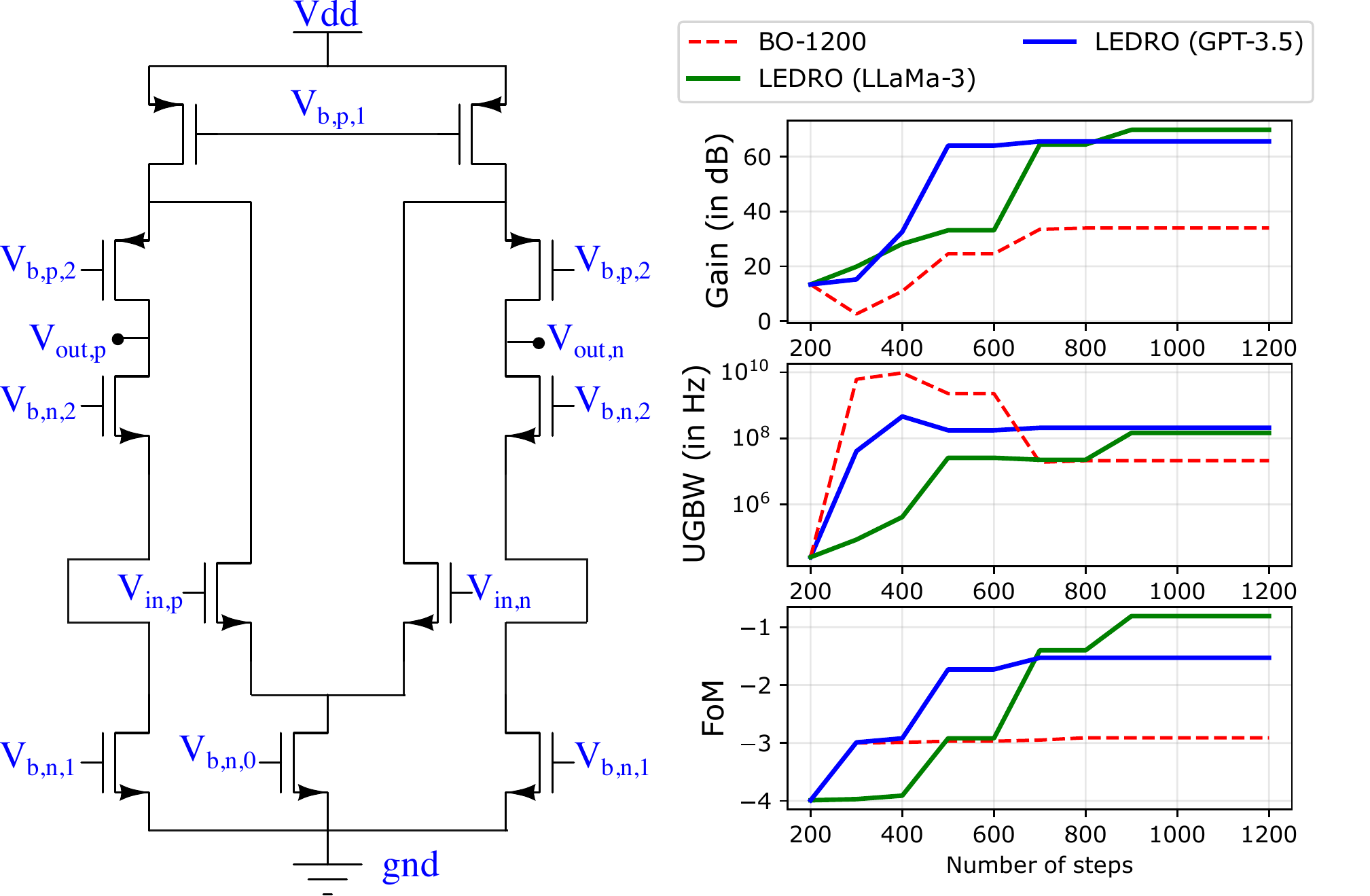}}
\vspace{-5pt}
\caption{Differential folded cascode circuit and gain, UGBW, and FoM for LEDRO with LLaMa3-70b, GPT-3.5-turbo models and BO-1200 vs steps.}
\label{fig6}
\vspace{-10pt}
\end{figure}


\subsection{Case Study: End-to-end example of 7nm differential folded cascode amplifier}
\label{sec:case-study}

We present a case study for the design of a differential-ended folded cascode circuit (as shown in Fig.~\ref{fig6}) in 7nm PTM-MG technology \cite{ptm} using LEDRO.
This circuit, chosen for its relevance in past literature \cite{fc1, fc2, fc3}, serves as a good benchmark for showcasing our model performance.
We conduct calibration point synthesis with TuRBO and add them to the first round prompt for the LLM as shown in Fig~\ref{fig7}.
For subsequent rounds, we follow up with optimizer feedback and LLM's self-reflection, as shown in Fig~\ref{fig8}.
We compare the optimization performance of two different runs of LEDRO utilizing two different LLMs with a pure TuRBO optimizer in terms of Gain, Unity Gain Bandwidth (UGBW), and FoM vs. optimizer steps in Fig~\ref{fig6}.
As observed, pure TuRBO over-optimizes UGBW and is unable to improve gain over the fixed 1200 iterations.
Although TuRBO might be able to reach a similar optimal point, it would take many more optimization steps.
Both runs of LEDRO achieve stronger gain, UGBW, and, in turn, high FoM in fewer optimizer steps.
We conduct future comprehensive experiments using only LLaMa3-70B \cite{llama3} since we validated the utility of GPT-3.5-turbo through this case study and it is more expensive to run the GPT model.

\begin{figure}[t]
\centerline{\includegraphics[scale=0.51]{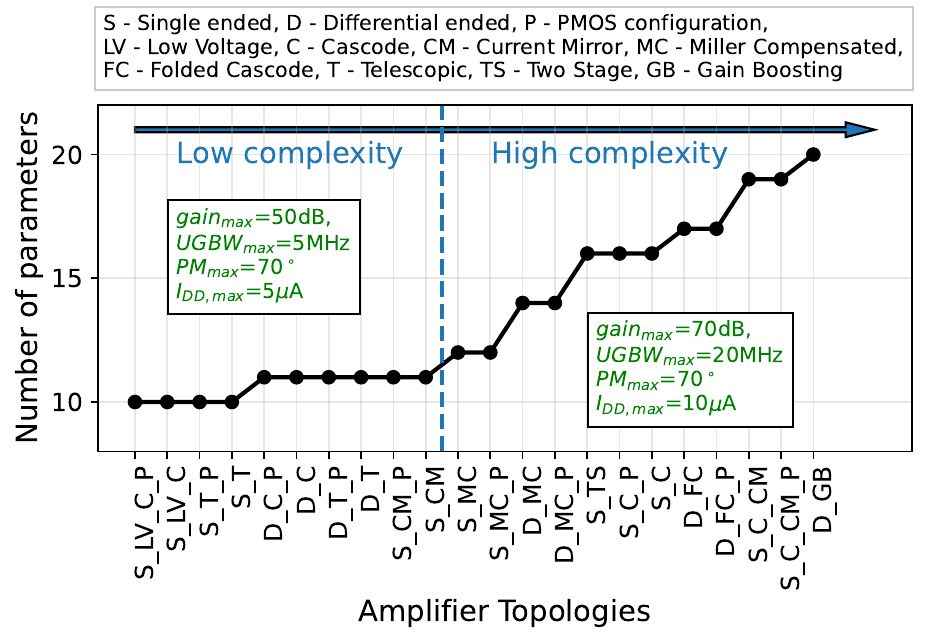}}
\vspace{-5pt}
\caption{Parameters per Op-Amp and division into high and low-complexity Op-Amps and their boundary values.}
\label{fig2}
\vspace{-10pt}
\end{figure}

\begin{figure*}[t]
    \centering
    \includegraphics[width=0.9\textwidth]{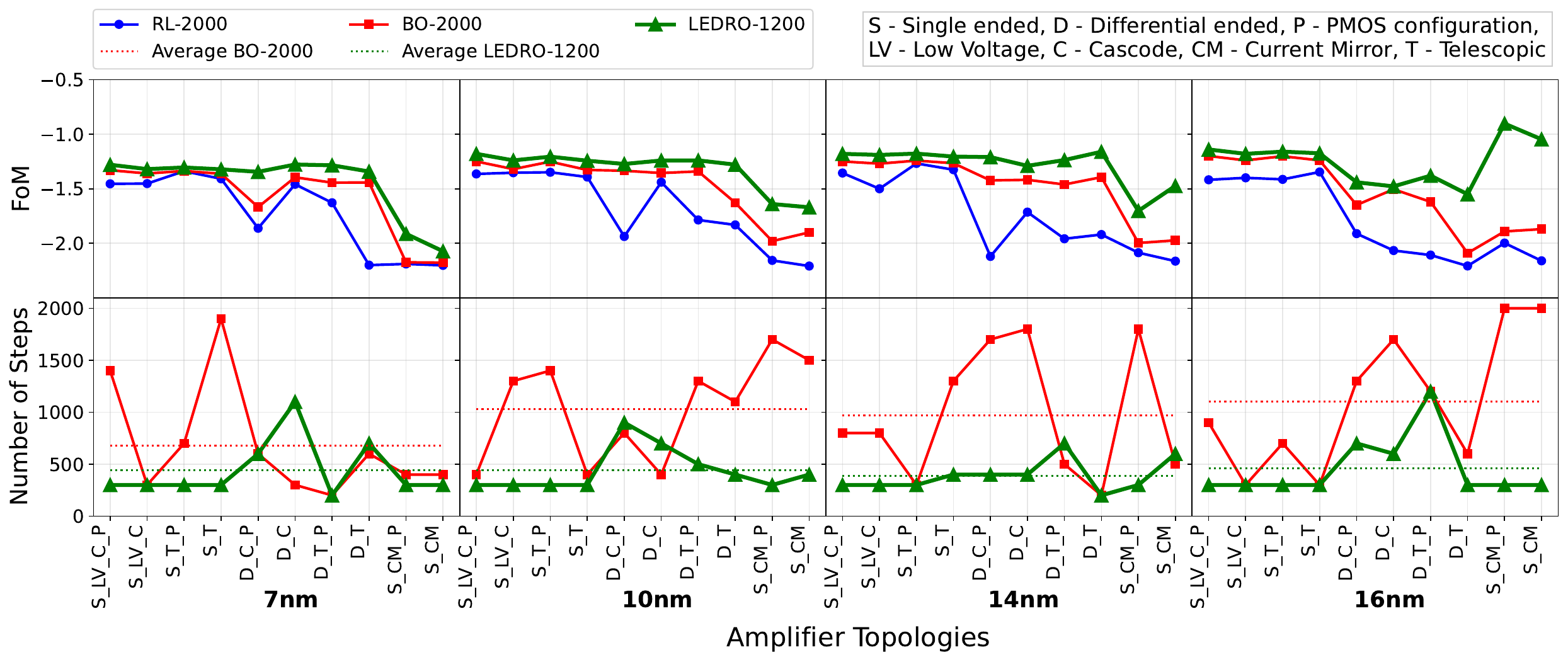}
    \vspace{-10pt}
    \caption{Performance (FoM) and Efficiency (number of steps) comparison of LEDRO with other baselines for different low-complexity amplifier topologies across four different technology nodes. As shown, LEDRO not only achieves better FOM, but also takes fewer steps to reach the best BO-2000 FoM.}
    \label{fig3}
    \vspace{-10pt}
\end{figure*}

\section{Experiments and Results}
\label{sec:experiments}
We describe our experimental setup in \ref{sec:experimentation}, the main results in \ref{sec:results} and supporting analysis in \ref{sec:analysis}.

\subsection{Experimental Setup}
\label{sec:experimentation}

\noindent\textbf{{Benchmarks:}}
To demonstrate the broad generalizability of our framework, we conduct experiments using a diverse set of 22 amplifier topologies, including both PMOS and NMOS input configurations with single-ended and differential-ended output designs, ranging from 7 to 19 transistors.
To create this benchmark, we modify and repurpose the subcircuit annotation task in \cite{gana} to design the parametrized FinFET netlists.
For richer benchmarking, we split the amplifiers into 10 low-complexity and 12 high-complexity Op-Amps based on the number of optimization parameters.
To ensure robustness, we utilize predictive technology models for multi-gate transistors (PTM-MG) \cite{ptm} across multiple technology nodes, specifically 7nm, 10nm, 14nm, and 16nm.
We show the different Op-Amps with their categorization and number of parameters in Fig~\ref{fig2}.

\noindent \textbf{{Evaluation Metric:}}
We consider performance and efficiency as the major evaluation metrics.
Following previous works like \cite{ab3, tbss}, we utilize the number of optimization steps to measure the efficiency.
For performance, the evaluation metric is the Figure of Merit (FoM), as we defined in ~\ref{sec:calibration-point-synthesis}, using user-defined specification boundary values.
Different amplifiers have different characteristics - some achieve higher gain, some are optimized for low power, etc - and thus can have varied specification boundaries.
These characteristics can vary significantly when scaling the same topology across technology nodes.
Most previous works \cite{fc3, RL2} experiment with just 1-3 Op-Amps, which allows them to set customized FoM boundary values for each circuit easily.
Since our work focuses on a comprehensive benchmark of 22 Op-Amps across four technology nodes, it is challenging and infeasible to set customized boundary values for each circuit.

Instead, we specify two sets of boundary values - one each for the low-complexity and the high-complexity Op-Amp groups.
Specifically, for our low-complexity group, we set $G_{bound}$ = 50 dB, $I_{bound}$ = 5 {\textmu}A, $f_{bound}$ = 5 MHz, $PM_{bound}$ = 70{\textdegree}. For our high-complexity group, we set $G_{bound}$ = 70 dB, $I_{bound}$ = 10 {\textmu}A, $f_{bound}$ = 20 MHz, $PM_{bound}$ = 70{\textdegree}.
We set $w_G=3$, $w_f=w_{PM}=1$, and $w_I=-1$ for the FoM objective function.
These parameters are specifically chosen by circuit experts to be challenging for most amplifiers to rigorously evaluate the capabilities of the optimization algorithms.

\indent\textbf{{Baselines and Implementation:}}
We utilize Cadence Spectre to run our simulations.
Primarily, we compare our approach against the pure TuRBO optimizer \cite{turbo} which is proposed for automated analog sizing in \cite{turbo1}, and also used for subspace sampling in \cite{tbss}. TuRBO is run for 1200 steps (BO-1200), matching our method's maximum allowable evaluation budget (200 steps for calibration point synthesis and 100 optimizer steps $\times$ 10 LEDRO iterations).
As a stronger baseline, we also compare with an over-optimized TuRBO run with an evaluation budget of 2000 steps (namely BO-2000).
For a fair comparison, we ensure that LEDRO and the TuRBO baseline use the same starting points by continuing to train TuRBO using the initial 200 steps used for calibration point synthesis in LEDRO, and report the best of five runs.
We also include a Reinforcement Learning (RL) baseline, which was also run for 1200 and 2000 steps (namely RL-1200 and RL-2000).
For this baseline, we modify the base model from \cite{autockt}, with the recent improvements of actor-critic networks of \cite{RL1, RL2}.
We provide the actor with the region of operation knowledge and the critic with additional $v_{gs}$, $v_{ds}$, $g_m$, and $i_{ds}$ information for better optimization. 
Simulation-based approaches like genetic algorithms \cite{genetic1, genetic2} are sample-inefficient and poor in performance compared to TuRBO as shown in \cite{turbo1, autockt}.
AmpAgent’s\cite{ab1} device sizing agent primarily uses TuRBO algorithm. Since its code is not public, initial-value estimates were not available. Our method will show clear improvements over TuRBO, benchmarking against AmpAgent’s approach to our best capability. Despite efforts to replicate ADO-LLM’s \cite{ab3} process on our case study, it yielded suboptimal results. Due to unavailable implementation details, we do not include this baseline for our comparisons.

\begin{table}[]
    \centering
    \caption{Average figure of merit and optimization steps for low-complexity and high-complexity Op-Amps.}
    \setlength{\tabcolsep}{3.8pt}
    \begin{tabular}{@{} l|*{4}{c}|*{4}{c} @{}}
    \toprule
     & \multicolumn{4}{c@{}|}{Low-Complexity} & \multicolumn{4}{c@{}}{High-Complexity} \\  \cmidrule(l){2-9}
    Algorithm & 7nm & 10nm & 14nm & 16nm & 7nm & 10nm & 14nm & 16nm \\ 
    \midrule
    \multicolumn{9}{c@{}}{\textsc{Avg Figure of Merit (FoM)  $\mathrm{\uparrow}$}} \\
    \midrule
      RL-1200  & -1.79 & -1.73 & -1.76 & -1.82 & -2.69 & -2.64 & -2.27	& -2.24 \\
      RL-2000  & -1.72 & -1.68 & -1.74 & -1.80 & -2.62 & -2.55 & -2.20 & -2.13 \\
      BO-1200  & -1.57 & -1.48 & -1.57 & -1.61 & -1.46 & -1.26 & -1.18 & -1.16 \\
      BO-2000 & -1.57 & -1.47 & -1.47 & -1.55 & -1.19 & -1.21 &-1.12 & -1.08 \\
       \midrule
      \textbf{LEDRO} & \textbf{-1.45} & \textbf{-1.32} & \textbf{-1.28} & \textbf{-1.25} & \textbf{-0.53} & \textbf{-0.62} & \textbf{-0.61} & \textbf{-0.64} \\
     \% boost & \multirow{ 2}{*}{8\%} & \multirow{ 2}{*}{10\%} & \multirow{ 2}{*}{13\%} & \multirow{ 2}{*}{20\%} & \multirow{ 2}{*}{55\%} &  \multirow{ 2}{*}{48\%} &  \multirow{ 2}{*}{45\%} &  \multirow{ 2}{*}{41\%} \\
     over BO-2000 & & & & & & & & \\
      \midrule
      \multicolumn{9}{c@{}}{\textsc{Avg number of steps to best BO-2000 FoM  $\mathrm{\downarrow}$}} \\
      \midrule
      BO-2000 & 680 & 1030 & 970 & 1040 & 992 & 1217 & 1042 &	875 \\
      \textbf{LEDRO} & 440 & 440 & 390 & 460 & 417 & 675	& 575 & 750 \\
    \bottomrule
    \end{tabular}
    \label{tab1}
    \vspace{-10pt}
\end{table}

\begin{figure*}[!ht]
    \centering
    \includegraphics[width=0.9\textwidth]{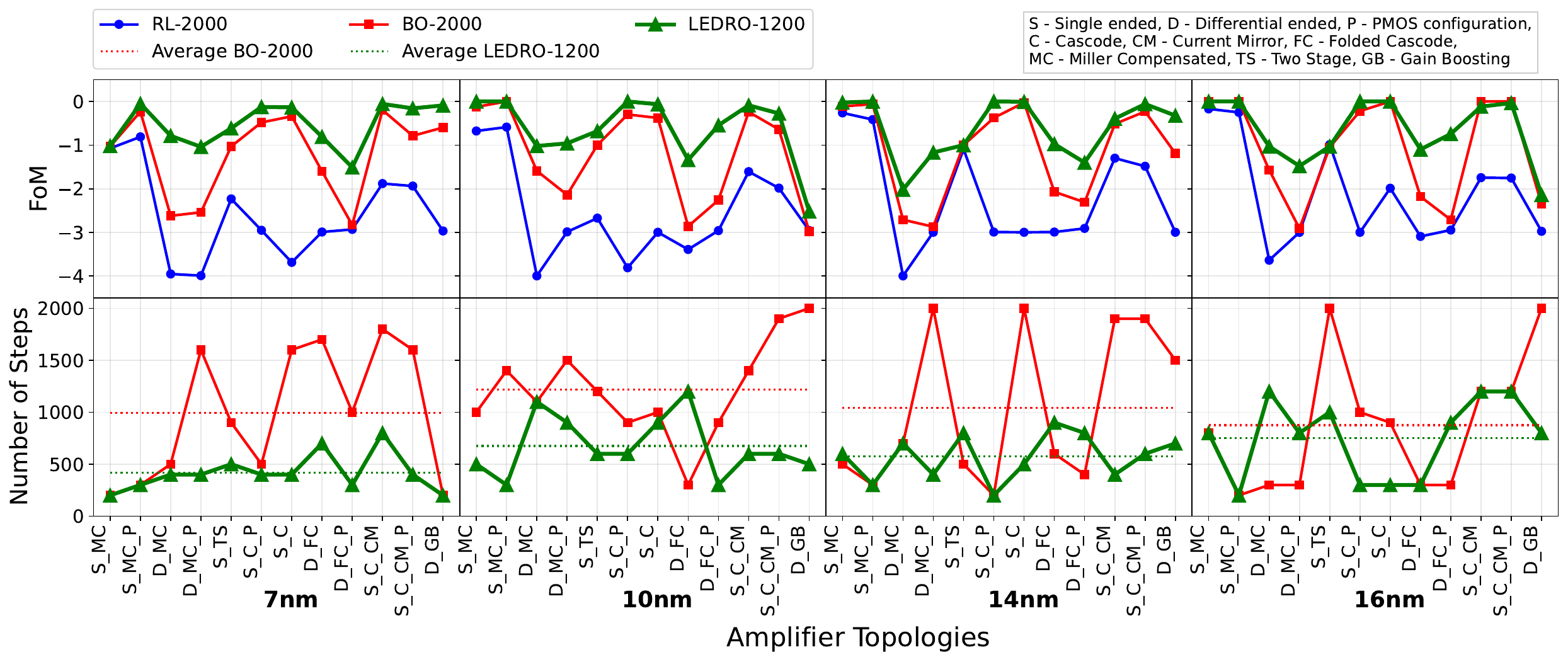}
    \vspace{-10pt}
    \caption{Performance (FoM) and Efficiency (number of steps) comparison of LEDRO with other baselines for different high-complexity amplifier topologies across four different technology nodes. As shown, LEDRO not only achieves better FOM, but also takes fewer steps to reach the best BO-2000 FoM.}
    \label{fig4}
    \vspace{-11pt}
\end{figure*}

\begin{figure}[h]
\centerline{\includegraphics[scale=0.4]{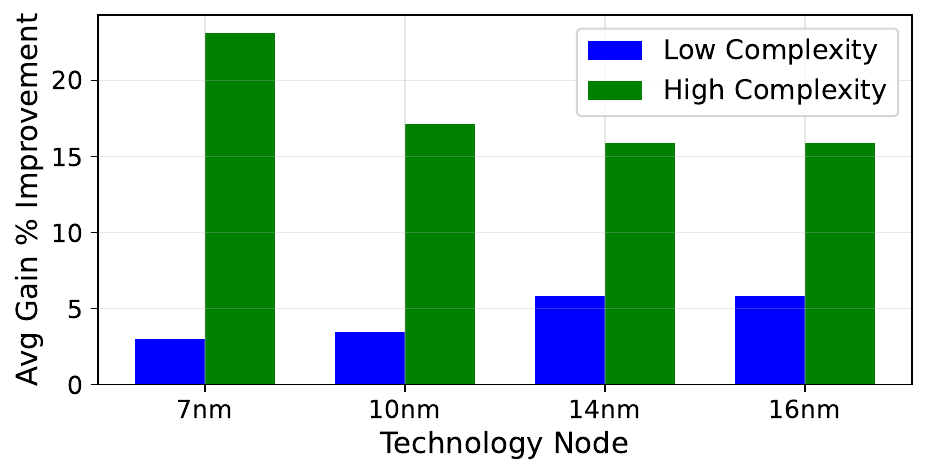}}
\caption{Average percentage improvement in gain for both sets of amplifiers across technology nodes.}
\label{fig5}
\vspace{-15pt}
\end{figure}

\subsection{Results}
\label{sec:results}
We show a detailed breakdown of the performance and efficiency for each amplifier topology and technology node in Fig~\ref{fig3} and Fig~\ref{fig4} for the low-complexity and high-complexity groups, respectively.
We notice a clear trend wherein our model LEDRO not only achieves the best FoM performance but also reaches optimal points in lesser number of steps.
Through Table~\ref{tab1}, we present a summary of the average maximum FoM achieved across the low and high-complexity groups of amplifier designs across the different technology nodes.
Our results show the 8-20\% and 16-21\% improvement in amplifier performance for low-complexity Op-Amps compared to the 2000-step TuRBO and 2000-step RL optimization, respectively, across technology nodes.
These improvements are further amplified in the high-complexity Op-Amps with 41-55\% improvement over TuRBO-2000 and 70-80\% over RL-2000. Notably, this increased improvement from low to high-complexity is consistently observed across all technology nodes. Such consistency suggests that while low-complexity and low-dimensional circuits are easy to optimize even with more simplistic methods, designing higher-complexity circuits is still challenging for conventional optimization algorithms.
On the other hand, our method LEDRO leverages LLMs to reduce and refine the high-complexity search space and efficiently use optimization on this lower-complexity refined space to provide strong improvements for the high-complexity circuits as well.
Such results highlight the robustness, generalizability, and scalability of LEDRO.

We also compare the optimization speed in terms of number of steps required by LEDRO and the best baseline BO-2000 to achieve the best FoM achieved by the BO-2000 algorithm, as shown in Table~\ref{tab1}.
Results show that LEDRO achieves an average speed-up of 1.54 $\sim$ 2.48$\times$ across low-complexity Op-Amps and  1.17 $\sim$ 2.38$\times$ for high-complexity Op-Amps across the technology nodes relative to the BO-2000 baseline. To compare computational latencies of BO-2000 and LEDRO, we compare the average time to achieve the best FoM achieved by the BO-2000 algorithm, while also accounting for the LLM prompt-response time. Relative to BO-2000, LEDRO achieves this optimum in: 0.49 $\sim$ 0.72$\times$ time for low-complexity Op-Amps 0.47 $\sim$ 0.98$\times$ time for high-complexity Op-Amps, across technology nodes. This demonstrates the superior optimization efficiency of LEDRO compared to traditional optimization techniques.

\subsection{Analysis: Op-Amp gain ($G$) improvement}
\label{sec:analysis}
When we examine the detailed improvements introduced by LEDRO, it is remarkable that LEDRO not only enhances the overall FoM, but also demonstrates exceptional efficiency in improving the Op-Amp gain.
The average improvement in terms of gain achieved by LEDRO relative to BO-2000 is plotted in Fig.~\ref{fig5}. It is noteworthy that the gain improvement is particularly significant for high-complexity circuits across different technology nodes, highlighting the advantage of using LEDRO in design automation.
The gain values were capped at $G_{bound}$ if either of the methods met the boundary gain value, ensuring a fair comparison. 
Overall, LEDRO provides an average improvement of about 15-23\% on high-complexity Op-Amps and about 3-6\% on low-complexity Op-Amps.



\section{Conclusion}
\label{sec:conclusion}
In this paper, we introduce LEDRO, which leverages LLMs in synergy with optimization techniques to efficiently explore design space of Op-Amps.
Our method advances analog circuit design automation by iteratively refining the design search space through the LLM's prior knowledge and reasoning, thus facilitating strong optimization performance and better designs without human expertise. 
Experiments across a wide range of topologies and technology nodes demonstrate LEDRO's generalizability with 13-48\% FoM improvements and efficiency with 1.7 $\sim$ 2.15$\times$ speed-ups compared to conventional optimization techniques.
Furthermore, LEDRO is significantly stronger in designing high-complexity and high-dimensional circuits, which provides promise for more practical use cases.
In conclusion, the generalizability and adaptability with other optimization techniques marks an advancement in the field, offering a solution for complex analog circuit design challenges.

\section*{Acknowledgment}
We thank MIT Microsystems Technology Laboratory (MTL) for compute and simulation environment access, and IBM BAM for LLM API access.


\vspace{12pt}
\clearpage
\section{Appendix}
\subsection{Additional experiments and ablations}

\noindent \textbf{Temperature robustness:} We evaluated LEDRO's robustness across different temperature values, specifically for the case study in Sec.~\ref{sec:case-study}. The best performance for the case study was achieved at temperature = 0.8, which was used for all experiments for consistency. Other temperatures (0.5–0.9) yielded performances within 0.4 FoM of the best setting and still outperformed the baseline by $>$1.7 FoM. Higher temperatures produced more diverse search spaces, while lower temperatures led to more deterministic explorations.
The best temperature=0.8 is high, and it suggests that the added diversity due to sampling aids the LLM for better exploration for the optimization problem.

\noindent \textbf{LLM Generalizability:} While LLaMA3-70B \cite{llama3} was primarily used, LEDRO is not restricted to a single LLM. Results in Fig~\ref{fig6}. demonstrate its effectiveness with GPT-3.5, confirming its versatility. Due to resource limitations, we conduct future comprehensive experiments only using LLaMa3-70B since we validated the utility of GPT-3.5-turbo through the case study and also since it is more expensive to run the GPT model. 
To establish this generalizability across LLMs better, we conduct additional experiments with the newer LLaMA3.3-70B.
Our experiments revealed that LLaMA3.3-70B runs also show comparable performance.

\noindent \textbf{Ablating the feedback loop and reflection mechanism:}
To emphasize the importance of two components we use in our LLM pipeline, we conduct a small ablation study on the case study by removing these components individually and comparing the model performance.
First, removing the feedback loop reduces LEDRO to a single iteration, causing a 100–300\% FoM reduction.
This emphasizes its crucial role in iterative improvement, which lead to continual improvements.
Secondly, removing the self-reflection prompts results in a 120\% FoM reduction.
This highlights the significance of self-guided reflections in calibrating the model's outputs and in turn, refining the design space.

\noindent \textbf{Latency and complexity analysis:}
Since LLM-based and non-LLM approaches are vastly different, comparing their computational complexity is not trivial.
Nonetheless, we attempt to compare their complexities in terms of runtime computes.
Specifically, we consider the two best models LEDRO and BO-2000 and compare the average runtimes to reach the best optimum of BO-2000.
For BO-2000, we consider the time taken for the TuRBO algorithm to reach the optimum.
For LEDRO, we take the summation of the time for the LLM API calls (accounting for the LLM prompt response time) and the time to run the intermediate TuRBO iterations.
Overall, relative to BO-2000, LEDRO achieves this optimum 0.55$\times$ faster for low-complexity and 0.68$\times$ faster for high-complexity Op-Amps.
These gains stem from iterative design space narrowing, minimizing the number of optimization steps.

\subsection{Discussion and Comparison about recent LLM works in analog circuit sizing}
Recent times have seen a surge in the utilization of LLMs for analog circuit design. While we briefly cover relevant literature in the main text, this appendix provides additional clarification regarding the scope and applicability of comparison with selected works.

AnalogCoder\cite{ab2} primarily addresses functional verification rather than transistor-level optimization, as noted in their paper. Their objectives differ from ours, making a direct performance comparison not meaningful in this context.

AmpAgent\cite{ab1} employs the TuRBO\cite{turbo} algorithm as the core of its device sizing agent. While we aimed to evaluate our approach relative to AmpAgent, the absence of public code and initial-value estimates limited the reproducibility of their exact setup. Nonetheless, we benchmarked against the TuRBO algorithm, to the best of our ability, and observed clear improvements.

For ADO-LLM\cite{ab3}, we attempted to adapt their methodology to our case study described in Sec.~\ref{sec:case-study}. However, the results were suboptimal, likely due to limited implementation details. To ensure accuracy and fairness, we chose not to include these findings. It is also worth noting that, in contrast to ADO-LLM, our work provides extensive benchmarking against a wide range of recent RL and BO approaches.

\subsection{Further information}
We hypothesize that LLMs bring design expertise to optimization by identifying relationships in circuit netlists, like relative transistor-sizing. These insights generalize across technology nodes, avoiding computationally prohibitive optimization over the entire parameter space defined by technology model files. By narrowing the design space to high-potential regions, LLMs enable faster, more efficient searches, boosting performance. We would also like to emphasize that our method requires no LLM training or fine-tuning and just works with LLM prompting. 


While LLMs introduce decision-making complexity, this challenge also applies to RL/BO methods, which rely on opaque neural network models. Our framework improves design efficiency compared to traditional human-driven processes, which aligns with the increasing adoption of AI/LLMs across circuit design tasks. Our methodology allows for refinement based on circuit knowledge, mitigating concerns around interpretability while enhancing design convergence.

As for the scope of our framework, while our evaluations focus on Op-Amp designs, the proposed approach is generalizable to broader analog circuit design tasks. The framework's heuristic-driven design space reduction can be applied across multiple domains, including RF circuits, data converters, and other automation tasks where informed design space exploration accelerates convergence and efficiency. Future work aims to explore these extensions further.  
\end{document}